\documentclass[12pt, reqno]{llncs}
\usepackage{amsmath,amscd,amssymb}
\usepackage{graphicx}
\usepackage{subfigure}
\usepackage{dsfont}
\usepackage{xcolor}
\usepackage{multirow}
\usepackage{amsbsy}
\usepackage{appendix}
\usepackage{mathtools}
\usepackage[hyphens]{url}
\makeatletter
\renewcommand\paragraph{\@startsection{paragraph}{4}{\z@}%
            {-2.5ex\@plus -1ex \@minus -.25ex}%
            {1.25ex \@plus .25ex}%
            {\normalfont\normalsize\bfseries}}

\begin{document}

\title{Born for Auto-Tagging: Faster and better with new objective functions\thanks{We thank Tung-Chi Tsai and Liang-Wei
 Liu for the technical support and discussion during the preparation of the paper.}}

\titlerunning{Born for Auto-Tagging}

\author{Chiung-ju Liu \inst{1}\and Huang-Ting Shieh
\inst{1}}
\authorrunning{C. Liu \and H. Shieh}
\institute{awoo  Intelligence AI\\
\email{\{cjliu, timshieh\}@awoo.ai}}
\maketitle             
\begin{abstract}
Keyword extraction is a task of text mining. It is crucial to increase search volume in SEO and ads. Implemented in auto-tagging, it makes the process on a mass scale of online articles and photos efficiently and accurately. BAT was invented for auto-tagging, serving as awoo's AI marketing platform (AMP). It not only provides service as a customized recommender system but also increases the converting rate in E-commerce. The strength of BAT converges faster and better than other SOTA models, as its 4-layer structure achieves the best F scores at 50 epochs. In other words, it performs superior to other models which require deeper layers at 100 epochs. To generate rich and clean tags, we create new objective functions to simultaneously maintain similar ${\rm F_1}$ scores with cross-entropy while enhancing ${\rm F_2}$ scores. To assure the even better performance of F scores, we revamp the learning rate strategy proposed by Transformer~\cite{Transformer} to increase ${\rm F_1}$ and ${\rm F_2}$ scores at the same time.  
\end{abstract}

\section{Introduction}
Keywords are widely used for site search, ads, and personalized recommendations on E-commerce, news sites, streaming media, etc. In particular, keyword search dominates the performance in search engine optimization (SEO). Valuable words such as high search volume are used to write articles with strong content or target popular issues. Automatically keyword extraction is helpful in online business.

Keyword extraction is essential in text mining, information retrieval, and natural language processing (NLP). However, it is challenging when it is implemented on web data or industry documents. The difficulty is not only due to the scale of data but also noisy signals including unstructured sentences, emotional symbols, network language, etc. Manual labeling for keywords is time-consuming. The consistency of labeling differs from labelers and needs rules to supervise. Machine learning emerges to solve the problem. Unsupervised algorithms based on statistic models are known as TF-IDF, BM25, and YAKE~\cite{YAKE}. Graph-based models include TextRank~\cite{TextRank}, Rake~\cite{RAKE}, Singlerank~\cite{singlerank} etc. Supervised learning with neural networks has had various developments recently. Some approaches are based on ranking the matching scores of queries with documents~\cite{DSSM,DRMM,KNRM,reranker}. Others are performed as semantic analysis classification~\cite{keyBLSTM,keyBERT}, etc.

awoo is successful in utilizing valuable tags in E-commerce, retail stores, AdTech, SEO, retargeting service, and business intelligence. awoo AMP uses tags to help customers to find products efficiently and increase conversion rates. Tagging for product pages not only advantage in keyword search but also benefits online stores in managing products.  Most personal recommendation systems are based on collaborative filtering. It focuses either on items that one bought in the past or other people bought in common. However, personal interest shifts or relating to popular items may cause poor recommendations. Understanding products and focusing on the message of the items by tags may capture customers' intentions. We use A/B test to show that it increases  71\% search volume, 230\%  time on page, 200\% conversion rates, and 100\% per customer transaction on average. It does induce great shopping serendipity.

To generate tags for large E-commerce platforms, we investigate extracting tags automatically. Tags range over categories, goods, brands, and features, varying in different industries. They are complicated and technical to establish rules. However, we have extensive experience in managing tags due to our long-term business in SEO.

\begin{figure}[ht]
\centering
\includegraphics[width=0.9\textwidth]{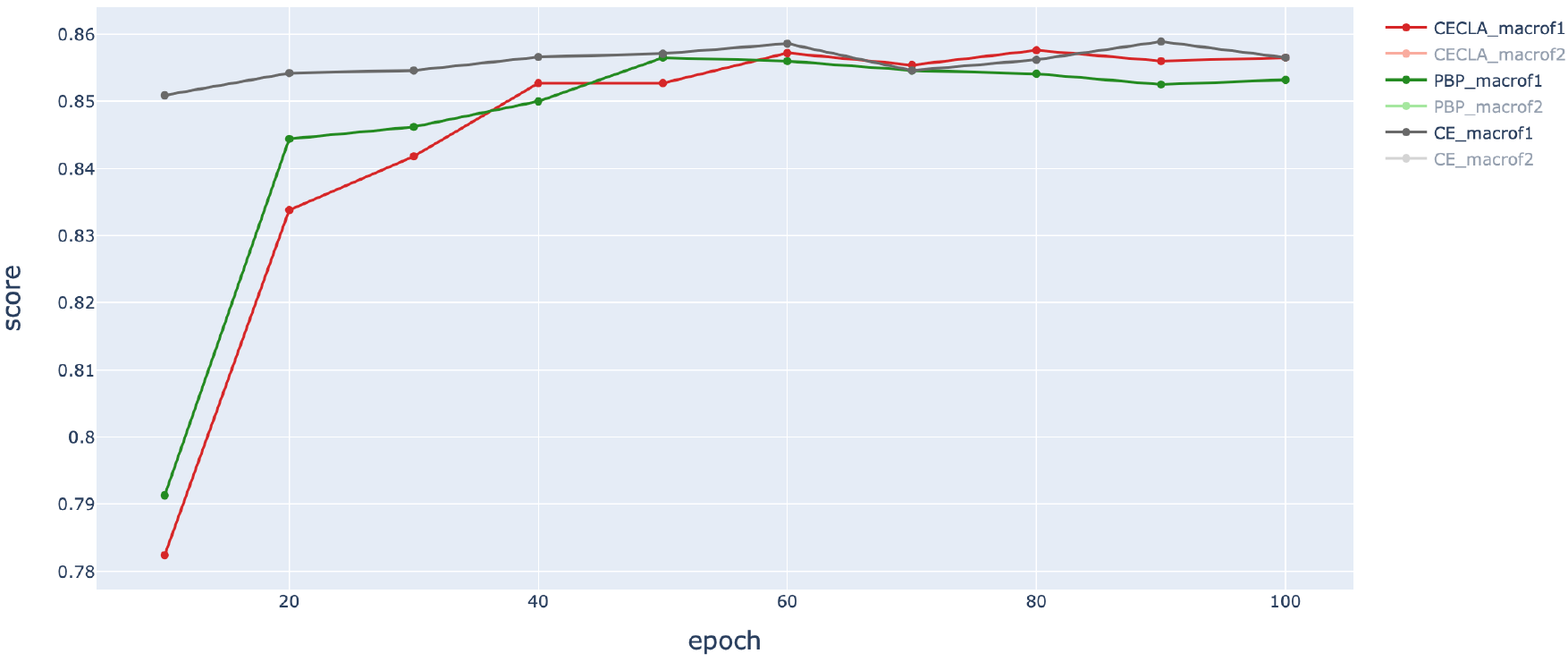} 
\caption{Loss comparison of CECLA, PBP, and CE with ${\rm F_1}$ scores}
\label{ceclapbpcef1}
\end{figure}

\begin{figure}[ht]
\centering
\includegraphics[width=0.8\textwidth]{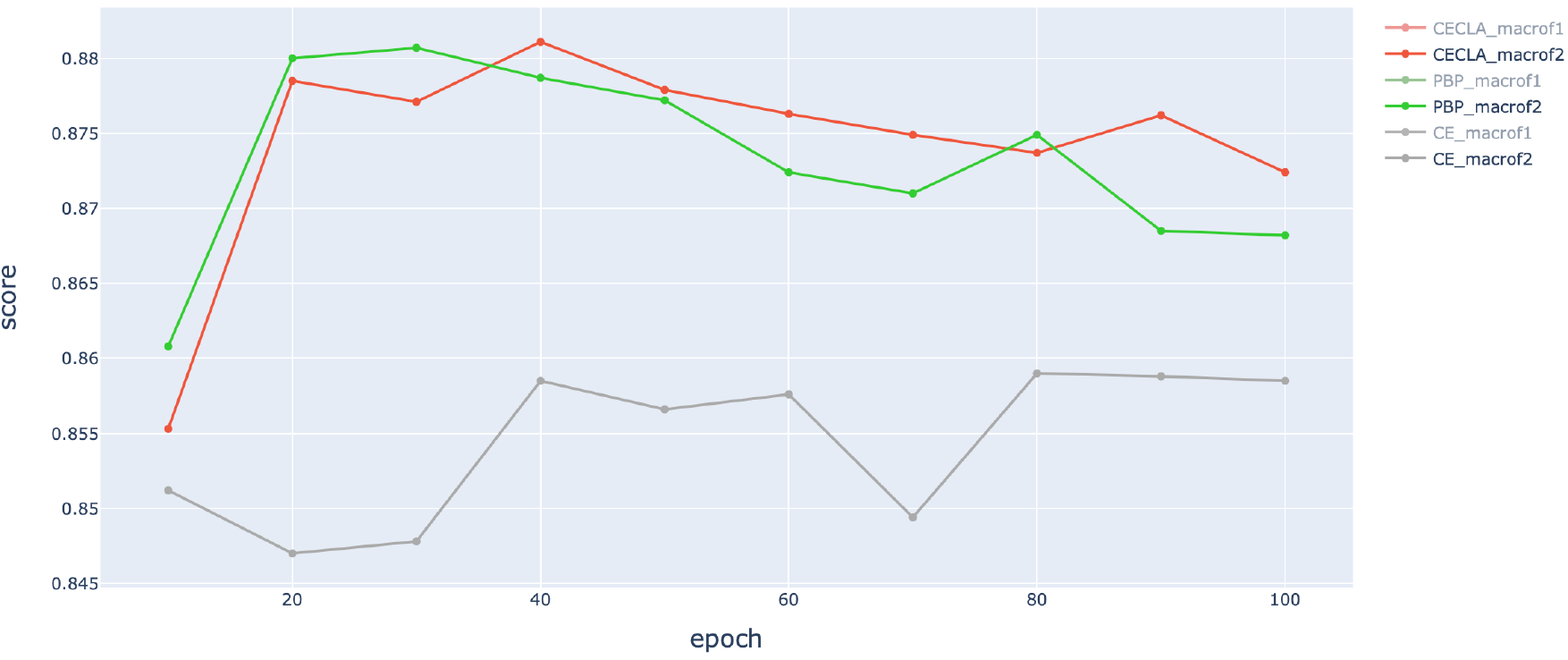} 
\caption{Loss comparison of CECLA, PBP, and CE with ${\rm F_2}$ scores}
\label{ceclapbpcef2}
\end{figure}

\begin{figure}[ht]
\center{
\includegraphics[scale=0.4]{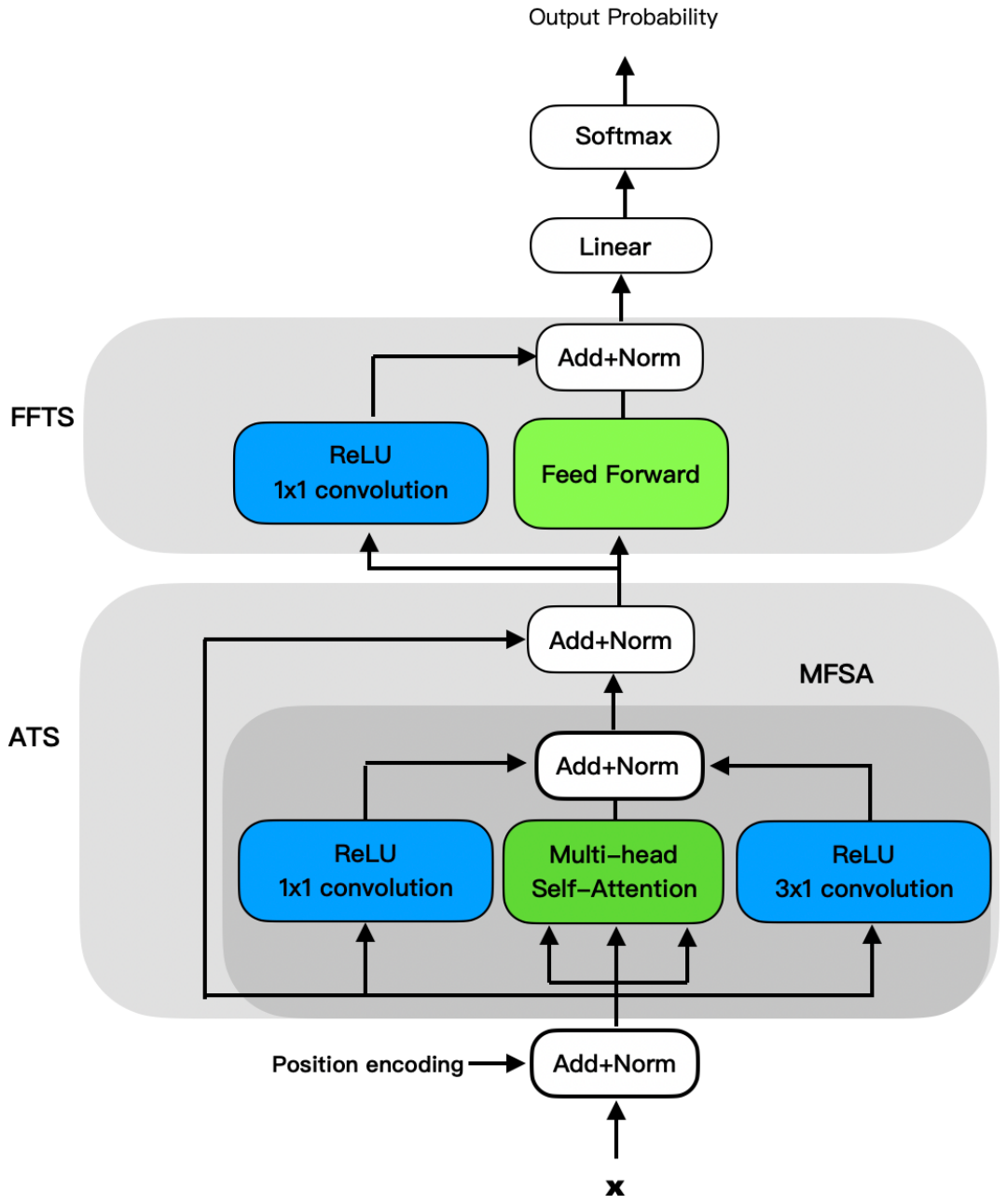} }
\caption{Architecture of BAT}
\label{en}
\end{figure}

BAT ({\bf B}orn for {\bf A}uto-{\bf T}agging) is reconstructed from the encoder of Evolved Transformer~\cite{ET}. This model is created for auto-tagging. It converges faster and better than other SOTA models. Its 4-layer structure (figure~\ref{en}) obtains the best F scores at 50 epochs and the scores exceed the other deep layers at 100 epochs. To generate rich (higher ${\rm F_2}$) and clean (higher ${\rm F_1}$) tags, we create new objective functions $L_{\rm PBP}$ \eqref{pbp} and $L_{\rm CECLA}$ \eqref{awooloss} to maintain similar ${\rm F_1}$ scores with cross-entropy $L_{\rm CE}$ but also increase ${\rm F_2}$ scores simultaneously. See figure \ref{ceclapbpcef1} and \ref{ceclapbpcef2} for detail. 

Industry aspects such as SEO and ads value ${\rm F_2}$ scores more than ${\rm F_1}$ scores.  On the other hand, security-related businesses such as fingerprint recognition or medical object detection are concerned about precision or ${\rm F_1}$ scores. Since we want to find not only plentiful but also neat tags,  we look for the relative maximum of macro ${\rm F_2}$ curves, not the lowest loss value. The optimal point may not occur at the end of the training by minimizing the loss. To overcome this problem and enhance a better convergence, we revamp the learning rate strategy \eqref{lr} proposed by Transformer~\cite{Transformer} to increase ${\rm F_1}$ and ${\rm F_2}$ scores at the same time. See table \ref{table1} for complete information.

In the classification task on EC-Zh data, BAT gets 1.7\%, 3.1\%, 3.3\% and 2.7\%  more than Transformer  via $L_{\rm CE}$, $L_{\rm WCE}$, $L_{\rm CECLA}$, and $L_{\rm PBP}$ objective functions with  respectively in macro ${\rm F_1}$ at 50 epochs. BAT obtains 2.9\%, 2.2\%, 2.3\% and 2.2\%  more than Transformer by $L_{\rm CE}$, $L_{\rm WCE}$,  $L_{\rm CECLA}$ and $L_{\rm PBP}$ loss functions with  respectively in macro ${\rm F_2}$ on EC-Zh data. Most SOTA models maintain almost the same ${\rm F_1}$ scores via new loss functions and gain 2\% ${\rm F_2}$ on average by comparing to $L_{\rm CE}$. Moreover, on EC-JP dataset, BAT gets 1.4\%, 2.5\%, 3.5\%, and 2.8\% in macro ${\rm F_1}$ via $L_{\rm CE}$, $L_{\rm WCE}$,  $L_{\rm CECLA}$ and $L_{\rm PBP}$ with  respectively. BAT gets 1.7\%, 0.8\%, 1.6\%, and 1.5\% in macro ${\rm F_2}$ via $L_{\rm CE}$, $L_{\rm WCE}$,  $L_{\rm CECLA}$ and $L_{\rm PBP}$ with  respectively. See the experiment of comparison in section \ref{sotacomparison}.

\section{Related Work}
\subsection{Contrastive Learning}
An objective function plays an important role in supervised learning. It directly  
 LeCun etc. introduced the contrastive loss function in~\cite{CL}. It is well known as triplet loss~\cite{facenet} in image classification.   Given a pair $\{x_+, x_-\}$ of a positive and negative sample and a training sample $x$, the object is to make $x$ look similar to $x_+$ and not similar to $x_-$ if $x$ is labeled as positive ($y=1$).  The triplet loss is formulated as 
 \begin{equation}
 L=\max(0, \|x-x_+\|^2-\|x-x_-\|^2+\alpha)\label{contrastivelearning}
 \end{equation}
 for binary classification. The model is supervised to learn to draw close to $x_+$ and push away from $x_-$ with a marginal distance $\alpha$. The feature space is measured in $L_2$-norm. It is applied to multiple classification~\cite{MCL}. The $(N+1)$-tuplet loss is to optimize identifying a positive sample from $N$ negative samples. It is 
 \begin{equation}
L(x_i,x_j)={\bf 1}\{y_i=y_j\}\|x_i-x_j\|^2+{\bf 1}\{y_i\neq y_j\}\max(0, \alpha-\|x_i-x_j\|^2)
 \end{equation}
When $N=2$, it is equivalent to triplet loss. Since it is not easy to train an NLP model by using $L_2$-norm, we prefer to use cross-entropy in binary classification:
\begin{equation}
L=y\log p+(1-y)\log(1-p),
\end{equation}
where $y=1$ if $x$ is positive and $y=0$ if $x$ is negative, and $p$ is the prediction of $x$. For multiple classifications, we formulate it as equation \ref{cecl}. We discuss the objective functions exhaustively in section \ref{trainingloss}.

\section{Model Architecture}
BAT is a model reconstructed from Evolved Transformer’s architecture~\cite{ET}.  We use the encoder for text classification. The complete encoder-decoder structure is designed for machine translation and other tasks matching different input and output sequence lengths.  Attention mechanisms in RNN models~\cite{seq2seq1} integrate the total inference of the input sequence with each output which accelerates the development of progress in NLP. Transformer adopts full attention and exempts from sequential input which makes a leap of development in deep learning. It dominates machine learning in all aspects. Many models are developed from it. 
\subsection{Input}
The encoder imports a sequence of tokens $\{w_1,\cdots,w_n\}$ with vector embedding $\{{\bf x}_1$, $\cdots,{\bf x}_n\}$ respectively. Since Transformer uses position encoding to replace the sequential input, the encoder can input the data $X=[{\bf x}_1,\cdots, {\bf x}_n]^T$ $\in\mathbb{R}^{n\times d_{e}}$ and output the probability $[{\bf p}_1,\cdots, {\bf p}_n]^T\in \mathbb{R}^{n\times C}$ for each token at the same time. 

Transformer~\cite{Transformer} adds up the position encoding and embedding vectors to be the feature representation. One can change it to a linear or affine linear combination for richer inputs. GPT2~\cite{GPT2} moves the layer normalization (LN) before the self-attention block and obtains a better result. Admin~\cite{Admin} shows that Pre-LN maintains steady gradient distribution through each layer in the decoder but does not affect much in the encoder.  We have an ablation experiment for a text classification task with a 4-layer encoder.  Pre-LN obtains a slice better result. 
\subsection{Encoder}
The encoder is composed of a stack of identical layers (4 layers for commercial requirements). Each layer consists of two sublayers: attention and feed-forward tournament selection sublayers (ATS and FFTS). The competition forms differently: MFSA races with constant residual connection and FFNN races with a weighted residual connection. The two parallel branches compete through backpropagation to update parameters for better performance. 

\subsubsection{ATS: Attention Tournament Selection Sublayer}
ATS sublayer contains a multiple-feature self-attention block (MFSA)  and a residual connection in parallel. Inspired by~\cite{ET}, we extract multiple features from the input: multi-head self-attention, $1\times 1$-convolution, and $3\times 1$-convolution in parallel. It can be formulated as 
\begin{equation}
{\rm MFSA}=f_{\rm LN}({\rm MultiHead}(Q,K,V)+f_{1\times 1{\rm conv}}(X)+f_{3\times 1{\rm conv}}(X)).\label{mfsa}
\end{equation}
\subsubsection{MFSA: Multiple Feature Self-Attention} MFSA Adds up Three Features:
\begin{itemize}
\item The Multi-head attention is directly followed from Transformer~\cite{Transformer}. 
\item The $1\times 1$-convolution followed by the activation ReLU captures the relation of oneself. The activation ReLU plays a role as a dropout. 
\item The $3\times 1$-convolution followed by the activation ReLU captures the relation between the current word with its context words. This represents a local feature. 
\end{itemize}
\subsubsection{FFTS: Feed-Forward Tournament Selection Sublayer} FFTS is created for the competition between two branches: FFNN and weighted residual connection. An ablation experiment shows that FFNN needs support from weighted res-connection to achieve prosperity. 
 \section{Training}
Suppose that training data $\{({\bf x}_i, {\bf y}_i)\}_{i=1}^N$ is a set of pairwise vectors, where ${\bf x}_i\in \mathbb{R}^{d_e}$ is the embedding of word $w_i$ and ${\bf y}_i \in \mathbb{R}^C$ is its label. Labels are defined by humans. 
\subsection{Training Loss}\label{trainingloss}
An objective function affects the performance of an ML model crucially.

We create two types of objective functions for imbalanced data in multiple classification tasks.
\begin{itemize}
\item 
The CECL (categorical Cross-Entropy with Contrastive Learning) loss is given by 
\begin{equation}
L_{\rm CECL}=-\left({\bf  y} \log {\bf p}+({\bf 1}-{\bf y})\log({\bf 1}-{\bf p})\right).\label{cecl}
\end{equation}
 Given a sample with label ${\bf y}=[1,0,0]$, suppose that it has two two predictions  ${\bf p}_1=[0.4, 0.3, 0.3]$ and ${\bf p}_2=[0.4, 0.5, 0.1]$. They obtain the same $L_{\rm CE}$ value but different 
\[
L_{\rm CECL}({\bf p}_1)=-\log(0.4)-\log(0.3)-\log(0.3)=0.71\]
 and 
\[
L_{\rm CECL}({\bf p}_2)=-\log(0.4)-\log(0.5)-\log(0.1)=0.74.\]
 The false prediction ${\bf p}_2$ obtains a larger loss value. It induces the model to learn more efficiently for a single token. Contrastive learning in $L_2$-norm~\eqref{contrastivelearning} separates the positive and negative samples with a certain margin. On the other hand, contrastive learning with cross-entropy does not reveal the physical meaning but the probability value that approaches the positive ($y=1$) or negative samples ($y=0$). See equation \eqref{posgadient} and~\eqref{neggradient} as the point of view of gradients in Appendix \ref{gradientofloss}.
 \item The PBP (Punish Bad Prediction) loss is given as 
\begin{equation}
L_{\rm PBP}=-\left({\bf  y} \log {\bf p}+({\bf 1}-{\bf y})\odot \hat{{\bf y}} \log({\bf 1}-{\bf p})\right), 
\end{equation}
 where $\hat{{\bf y}}$ the predicted class of ${\bf p}$, which is a one-hot vector. $L_{\rm PBP}$ produces more loss value if the model predicts an incorrect category. In previous example, we get $L_{\rm PBP}({\bf p}_1)=-\log 0.4=0.4$ and $L_{\rm PBP}({\bf p}_2)=-\log 0.4-\log 0.5=0.7$. It converges faster then $L_{\rm CE}$ and $L_{\rm CECL}$.
 \end{itemize}
 \subsection{Training Loss for Imbalanced Data}
 To implement $L_{\rm CECL}$ loss on imbalanced data for better performance on positive samples, we adapt weights $\boldsymbol{\alpha}$ and $\boldsymbol{\beta}$
\begin{equation}
L_{\rm WCECL}=-\frac{1}{N}\sum_{i=1}^N \bigg( \boldsymbol{\alpha}{\bf y}_i ({\bf 1}-{\bf p}_i)^\gamma  \log {\bf p}_i+ \boldsymbol{\beta}({\bf 1}-{\bf y}_i) {\bf p}_i^\gamma  \log ({\bf 1}-{\bf p}_i)\bigg).\label{awooloss}
\end{equation}
By leveraging the positive and negative samples having equal expectations, we manipulate EE weights
\begin{equation}
\begin{cases}
\begin{array}{ll}
\boldsymbol{\alpha}_{EE}&=(\frac{N}{N_1}, \cdots, \frac{N}{N_C})\\
\boldsymbol{\beta}_{EE}&=(\frac{N}{(N-N_1)}, \cdots, \frac{N}{(N-N_C)})
\end{array}
\end{cases},\label{eeweight}
\end{equation}
where $N_1$, $N_2$, $\cdots$, $N_C$ denote the number of samples of each category, and $C$ denotes the number of categories. 

To produce specific weights according to their category amounts of negative samples, we create array weights:
\[A=(A_{ij})=
\begin{cases}
\begin{array}{cl}
\frac{N}{N_i}&\text{if $i=j$}\\
\frac{N}{(C-1) N_j} &\text{if $i\neq j$}
\end{array}
\end{cases}.
\] 
The CECLA (Cross-Entropy by Contrastive Learning with Array weight) loss is given as $L_{\rm CECLA}$ for $\boldsymbol{\alpha}_i=A{\bf y}_i$ and $\boldsymbol{\beta}_i=A{\bf y}_i$ in \eqref{awooloss}. In particular, $A{\bf y}_i$ extracts the $j$-th column of $A$ if ${\bf y}_i$ is in class $j$. 
 $L_{\rm CECLA}$ induces fast convergence. It obtains its optimal (macro ${\rm F_1}=85.7$ and ${\rm F_2}=88.1$) with learning rate $\alpha=7/15$, $\beta=-1/30$, and $\lambda=1.001$ in \eqref{adjustlr} at 40 epochs . It increases accuracy not only for  BAT but also for other SOTA models.

 $L_{\rm PBP}$ with balanced weights is given by 
 \begin{equation}
L_{\rm PBP}=-\frac{1}{N}\sum_{i=1}^N \bigg( \boldsymbol{\alpha}  {\bf y}_i ({\bf 1}-{\bf  p}_i)^\gamma \log {\bf p}_i + \boldsymbol{\beta} 
(({\bf 1}-{\bf y})\odot \hat{{\bf y}}){\bf p}_i^\gamma \log({\bf 1}-{\bf p}_i) \bigg)
 ,\label{pbp}
\end{equation}
where ${\bf y}_i, {\bf p}_i,  \hat{{\bf y}}, \boldsymbol{\alpha},  \boldsymbol{\beta} \in \mathbb{R}^C$. BAT obtains macro ${\rm F_1}=85.5$ and ${\rm F_2}=88.0$ with $L_{\rm PBP_{1,8}}$ at 30 epochs for a learning rate $\alpha=6/11$ and $\lambda=0.999$ in \eqref{adjustlr}. $L_{\rm PBP}$ indeed saves training costs.

\subsection{Learning Rate Strategy}\label{blr}
Transformer adapts the learning rate according to the following formula~\cite{Transformer}:
\begin{equation}
 lrate=d_{\rm model}^{-0.5}\cdot \min({\rm step\_num}^{-1/2} ,{\rm step\_num}\cdot {\rm warmup\_steps}^{-1.5}),\label{lr}
\end{equation}
with ${\rm warmup\_step}=4000$. To change the convergence of the learning, we adjust the learning rate.

Let $x={\rm step\_num}$ and  $S={\rm warmup\_steps}$. Given $\alpha>0$, we adjust the learning rate 
\begin{equation}
 lrate=\lambda \cdot d_{\rm model}^{-1/2}\cdot \min( S^\beta \cdot x^{-\alpha} ,S^{-3/2}\cdot x),\label{adjustlr}
\end{equation}
by finding a $\beta$ to satisfy the continuous property
\[
S^\beta\cdot x^{-\alpha}=S^{-3/2}\cdot x.
\]
In general, a smaller learning rate induces a slower and finer convergence. A relative larger $\alpha$ in \eqref{adjustlr} leads to faster convergence and higher  ${\rm F_1}$ scores. 
Here are some examples for $L_{\rm CECLA_{1,20}}$ at 50 epochs.
\begin{table}[ht]
\centering
\begin{tabular}{ c| c|c|c|c|c|c}
\hline
\textrm{lrate}&v1&v2&v3&v3.1&v4.1&v5\\ 
 \hline
 ${\rm F_1}$ &85.3&85.5&\textbf{85.6}&85.3&85.4&85.0\\
 \hline 
  ${\rm F_2}$&87.8&87.7&87.6&\textbf{88.1}&87.9&87.3\\
 \hline
  \end{tabular}
  \caption{Learning Rate Comparison}
  \label{table1}
  \end{table}
  
v1 represents the original learning rate \eqref{lr}. v2 represents $\alpha=6/11$, $\beta=1/22$. v3 represents $\alpha=6/13$, $\beta=-1/22$ and v3.1 represents ${\rm v3\ast 0.999}$.
v4 represents $\alpha=7/15$, $\beta=-1/30$ and v4.1 represents for ${\rm v4\ast 1.001}$.
 v5 represents $\alpha=11/20$, $\beta=1/20$.  
 
 Roughly speaking, v2 converges slower than v1 due to the learning rate power of $\frac{6}{11}<\frac12$. On the other hand, v3 converges faster than v1. v3 at 50 epochs is almost the same as v1 at 60 epochs. ${\rm v3\ast 0.999}$ converges slower than v3 and has higher ${\rm F_2}$ scores. Although v4 converges slower than v1, ${\rm v4\ast 1.001}$ converges faster than itself.  ${\rm v4\ast 1.001}$ obtains its optimal at 40 epochs with macro ${\rm F_1}=85.7$ and ${\rm F_2}=88.1$. 
 Although v5 does not perform well at 50 epochs, It obtains its optimal at 40 epochs with macro ${\rm F_1}=85.5$ and ${\rm F_2}=88.1$. 
 
 Each model suits a different learning rate due to its amount of parameters and architecture.

\section{Experiments}
\subsection{Training Data}
We have two training datasets. The EC dataset contains product pages which consist of titles and descriptions. We clean the data with a list of regulations and stop words.  

\begin{enumerate}
\item EC-Zh, which is collected from 5 E-commence platforms, includes cosmetics, furniture, fashion, shoes, and large retail store. Product pages are written in traditional Chinese and writing styles vary from platform to platform. It contains 2000 product pages from each platform and is divided into 80\% for training and 20\% for testing. The average length is 46.6 words per page and the longest page is 368 in the training set. This dataset has high variance and is challenging for text classification. 
\item EC-JP, which is collected from 11 E-commence platforms, includes fashion, interior decoration, farm produce, and multiple categories. Product pages are written in Japanese and have different writing styles. We remove page lengths over 400 due to the efficiency of model training. EC-JP is twice of EC-Zh. The average length is 30 words per page. 
\end{enumerate}
Due to the privacy policy of these two datasets, we are not able to release them to the public for free use. To verify the effectiveness of our model, we provide the sample code with CoNLL 2003 as the source. 

\subsection{Accuracy Comparison with SOTA}\label{sotacomparison}
\subsubsection{Hardware}
\begin{itemize}
\item 1080 GPU has 8G memory and its operating system is Ubuntu18.04. 
\item 1080Ti GPU has 11G memory and its operating system is Ubuntu21.10. 
\end{itemize}
The framework deployed on them depends on the official codes. MUSE, Admin, and Informer use PyTorch. Transformer, Switch, Performers, and BAT use TensorFlow.
Further setup for each model will be mentioned in the additional appendix.   
\subsubsection{Optimizer}
All experiments are trained with optimizer Adam~\cite{Adam} with $\beta_1=0.9$, $\beta_2=0.98$, and $\epsilon=10^{-9}$. We adapt the base learning rate \eqref{lr} with ${\rm warmup\_steps}=4000$. 

\subsubsection{Hyperparameters}
To maintain the efficiency and accuracy of the performance, we chose a relatively small model compared to Transformer. Here are the hyperparameters that we set for the architecture and experiments. 

\begin{table}[ht]
\centering
\begin{tabular}{c|c|c|c|c|c|c|c}
\hline
N&$d_{\rm model}$ &$d_{\rm ff}$&h&$d_{\rm k}$&$d_{\rm v}$& $p_{\rm drop}$&n\\
\hline
4&128&512&4&32&32&0&1.316 MM\\
\hline
\end{tabular}
\caption{Variations in the BAT architecture}
\label{hyperparameters}
\end{table}
$N$ represents the number of encoders and $n$ represents the total number of parameters. 
\subsubsection{Comparison on EC-Zh}

First, we compare the performance of the model BAT with Transformer~\cite{Transformer}, Switch Transformer~\cite{Switch}, Admin~\cite{Admin}, MUSE~\cite{MUSE}, Performer~\cite{Performers}, and Informer~\cite{Informer} in EC-Zh dataset with four loss functions. We use the official or open-source code from the third place for a fair comparison. We extract the encoder of these SOTA models and add a classification layer on top of it.  

The distribution of EC-Zh dataset is $N_1:N_2:N_3=21.2:1.9:1$.  All models are built with 4-layer, 4 heads, and no dropout.  The following experiments are trained under the same conditions: learning rate strategy, batch=8, and 50 epochs.  The accuracy comparison of macro $\textrm{F}_1$ and $\textrm{F}_2$ are displayed in table \ref{f1sotacomparison} and \ref{f2sotacomparison} with respectively. 

\begin{table}[ht]
\centering
\begin{tabular}{c|c|c|c|c|c|c|c}
\hline
loss$\diagdown$ Model&BAT&Admin&Switch&MUSE&Transformer&Performers&Informer\\
\hline
CE&{\bf 85.7}&84.5&82.1&84.2&84.0&83.9&84.6\\
\hline
WCE&{\bf 85.5}&84.5&79.6&81.2&82.9&83.0&80.8\\
\hline
${\rm PBP}_{0,8}$&{\bf 85.6}&84.0&79.1&84.0&82.9&82.7&81.2\\
\hline
${\rm CLA}_{1, 20}$&{\bf 85.3}&84.3&75.0&83.7&82.4&82.1&80.9\\
\hline
\end{tabular}
\caption{${\rm F_1}$ SOTA comparison on EC-Zh}
\label{f1sotacomparison}
\end{table}

\begin{table}[ht]
\centering
\begin{tabular}{c|c|c|c|c|c|c|c}
\hline
loss$\diagdown$Model&BAT&Admin&Switch&MUSE&Transformer&Performers&Informer\\
\hline
CE&{\bf 85.7}&83.7&81.8&83.4&83.8&82.8&84.0\\
\hline
WCE&{\bf 87.9}&86.2&84.6&87.2&85.5&86.4&85.6\\
\hline
${\rm PBP}_{0,8}$&{\bf 87.7}&86.4&84.5&87.6&85.5&86.4&86.0\\
\hline
${\rm CLA}_{1, 20}$&{\bf 87.8}&86.0&82.5&87.7&85.3&86.2&86.2\\
\hline
\end{tabular}
\caption{${\rm F_2}$ SOTA comparison on EC-Zh}
\label{f2sotacomparison}
\end{table}
BAT is a model that converges faster than other SOTA models for the overall performance.  BAT obtains the best F scores at 50 epochs and is better than the deeper structure of other SOTA models at 100 epochs. ${\rm CLA}_{1, 20}$ represents CECLA loss with focal factor $\gamma=1$ and multiplication factor $\lambda=20$ in \eqref{lambda}. ${\rm PBP}_{0,8}$ stands for PBP loss with EE weight, focal factor $\gamma=0$ and $\lambda=8$ in \eqref{lambda}. 
\subsubsection{Comparison on EC-JP dataset}
The distribution in the EC-JP dataset is $N_1:N_2:N_3=9.57:1:1.01$.  All models are built with 4-layer, 4 heads, and no dropout.  The following experiments are all trained under the same conditions: Adam optimizer, learning rate strategy \eqref{lr}, and batch=8.  Table \ref{f1JPsotacomparison} is the accuracy comparison of SOTA models on macro ${\rm F_1}$ . Table \ref{f2JPsotacomparison} shows the accuracy comparison of SOTA models on macro ${\rm F_2}$. 
\begin{table}[ht]
\centering
\begin{tabular}{c|c|c|c|c|c}
\hline
loss$\diagdown$Model&BAT&Admin&MUSE&Transformer&Performers\\
\hline
CE&{\bf 85.8}&84.8&84.7&84.4&84.8\\
\hline
WCE&{\bf 85.4}&84.6&84.8&83.4&81.2\\
\hline
${\rm PBP}_{0,8}$&{\bf 85.6}&83.4&82.5&83.6&80.3\\
\hline
${\rm CLA}_{1, 20}$&{\bf 85.6}&83.9&82.6&82.1&80.9\\
\hline
\end{tabular}
\caption{${\rm F_1}$ SOTA comparison on EC-JP}
\label{f1JPsotacomparison}
\end{table}

\begin{table}[ht]
\centering
\begin{tabular}{c|c|c|c|c|c}
\hline
loss$\diagdown$Model&BAT&Admin&MUSE&Transformer&Performers\\
\hline
CE&{\bf 86.1}&84.2&85.1&84.4&85.6\\
\hline
WCE&{\bf 88.3}&87.7&87.3&87.5&86.5\\
\hline
${\rm PBP}_{0,8}$&{\bf 88.6}&87.3&87.6&87.2&87.2\\
\hline
${\rm CLA}_{1, 20}$&{\bf 88.3}&87.5&87.5&86.7&86.3\\
\hline
\end{tabular}
\caption{${\rm F_2}$ SOTA comparison on EC-JP}
\label{f2JPsotacomparison}
\end{table}
We adopt the learning rate \eqref{lr} by multiplying 0.93 for CE loss to slow down the convergence. For CECLA loss, we adopts $\alpha=6/11$ and $\beta=1/22$ for \eqref{adjustlr} to increase $\textrm{F}_2$ score. Since the data of EC-JP is twice of EC-Zh and contains more positive samples, most SOTA  models have higher F scores on EC-JP.

\subsubsection{Layers Comparison on EC-Zh} 
Experiments are done with the same conditions: ${\rm CLA}_{1, 20}$ loss function, batch = 8, 100 epochs. Increasing training time is concerned with the convergence of deeper layers. Table \ref{f1layercomparison} demonstrates the comparison of macro ${\rm F_1}$. Table \ref{f2layercomparison} demonstrates the comparison of macro ${\rm F_2}$.
\begin{table}[ht]
\centering
\begin{tabular}{c|c|c|c|c|c|c|c}
\hline
layer$\diagdown$Model&BAT&Admin&Switch&MUSE&Transformer&Performers&Informer\\
\hline
4-layer &{\bf 85.7}&83.9&79.1&84.4&83.0&82.7&82.8\\
\hline
6-layer &{\bf 85.4}&82.9&76.1&84.2&82.9&83.7&83..6\\
\hline
12-layer &{\bf 85.5}&83.6&80.2&85.4&82.1&82.9&83.4\\
\hline
24-layer &85.1&83.4&71.0&{\bf 85.2}&33.9&84.3&82.3\\
\hline
\end{tabular}
\caption{${\rm F_1}$ layers comparison}
\label{f1layercomparison}
\end{table}

\begin{table}[ht]
\centering
\begin{tabular}{c|c|c|c|c|c|c|c}
\hline
layer$\diagdown$Model&BAT&Admin&Switch&MUSE&Transformer&Performers&Informer\\
\hline
4-layer &{\bf 87.2}&85.5&84.1&86.9&85.3&85.3&86.5\\
\hline
6-layer &{\bf 87.1}&85.5&82.7&86.7&84.2&85.3&86.4\\
\hline
12-layer &{\bf 86.9}&85.5&85.1&86.7&85.3&86.0&86.3\\
\hline
24-layer &{\bf 87.4}&86.2&78.3&86.2&37.1&87.0&86.5\\
\hline
\end{tabular}
\caption{${\rm F_2}$ layers comparison}
\label{f2layercomparison}
\end{table}
 Admin adds the initialization parameters and improves the performance in deeper layers. Switch and Performers maintain stable performance among different layers of models. MUSE especially has higher ${\rm F_1}$ scores in 12 and 24-layer structures. BAT is less sufficient in deeper layers. Transformer does not converge in the 24-layer structure with several versions of the learning rate. 
 \appendix
\section{Gradient of Loss Functions}\label{gradientofloss}
\subsection{Cross-Entropy with Contrastive Learning}\label{gradientofCECL}
The loss of weighted focal categorical cross-entropy with contrastive learning is given by
\begin{equation}
 L_{\rm WFCECL}=-\frac1N\sum_{i=1}^N(\boldsymbol{\alpha}{\bf  y}_i ({\bf 1}-{\bf p}_i)^\gamma\log {\bf p}_i+\boldsymbol{\beta}({\bf 1}-{\bf y}_i){\bf p}_i^\gamma \log ({\bf 1}-{\bf p}_i)), 
 \end{equation}
 where $\boldsymbol{\alpha}, \boldsymbol{\beta}, {\bf y}_i, {\bf p}_i \in \mathbb{R}^C$ and $C$ is number of categories. If we calculate the gradient for a  single token
  for which ${\bf y}$ is in class $c$ and 
   $\boldsymbol{\alpha}={\bf 1}$ and $\boldsymbol{\beta}={\bf 1}$, then we get
   \begin{equation}
   \frac{\partial L}{\partial p_c}=\gamma (1-p_c)^{\gamma-1}\log p_c -\frac{(1-p_c)^\gamma}{p_c}<0.\label{posgadient}
   \end{equation}
 For the other components $j\neq c$,  we get
   \begin{equation}
   \frac{\partial L}{\partial p_j}=- \gamma p_j^{\gamma-1} \log (1-p_j)+  \frac{p_j^\gamma}{1-p_j}>0. \label{neggradient}
   \end{equation}
   $L_{\rm CECL}$ supervises the model to increase $p_c$ in the correct class and decrease $p_j$ for $j\neq c$ by minimizing the loss value directly. It teaches the model to learn a single token faster.
 
 However, the backpropagation is performed per batch and the proportion of positive and negative samples in different categories varies from batch to batch. The distribution is also different from the whole dataset. We have done experiments with dynamic loss whose weights are re-calculated per batch. However, it learns not well from it. It may be easier for the model to learn from fixed patterns of mass behavior. 

Suppose that the amounts of each category in the dataset are $N_1:N_2:\cdots: N_C$ and $\sum_{i=1}^C N_i=N$. Furthermore, suppose that all predictions for positive and negative samples in each category are the same to simplify the analysis. We get
 \begin{equation}
 L=-\frac1N\sum_{j=1}^C \bigg(\alpha_j N_j (1-p_{j,+})^\gamma \log p_{j,+}+\beta_j (N-N_j) p_{j,-}^\gamma  \log (1-p_{j,-})\bigg),\label{positivenegative}
 \end{equation}
 where $p_{j,+}$ is the prediction for positive samples in class $j$ and $p_{j,-}$ is the prediction for negative samples for class $j$. 
  \begin{itemize}
 \item 
 For categorical cross-entropy $L_{\rm CE}$ ($\boldsymbol{\beta}={\bf 0}$, $\boldsymbol{\alpha}={\bf 1}$ and $\gamma=0$ in \eqref{positivenegative}), solve by Lagrange multiplier under the condition $p_1+p_2+\cdots+p_C=1$
 \begin{equation}
 \frac{\partial L}{\partial p_j}=-\frac{N_j}{N}\frac{1}{p_j}
 \end{equation}
 for $j=1,\cdots, C$. 
 If the data is very imbalanced, $L_{\rm CE}$ can not get optimal. For example: If $N_1=20 N_2$ and $N_1$ represent the numbers of negative samples, then $p_1=20 p_2=1$ and $p_2=0.05$ can’t get the correct class when $y_2=1$. Thus $L_{\rm CE}$ learns better on the major data. 
\item  For weighted cross-entropy $L_{\rm WCE}$ ( $\boldsymbol{\beta}={\bf 0}$ and $\gamma=0$), Lagrange Multiplier gives 
\begin{equation}
\frac{\partial L}{\partial p_j}=-\frac{\alpha_jN_j}{N} \frac{1}{p_j}.\label{wce}
 \end{equation}
It concludes that 
$\boldsymbol{\alpha}=(\frac{N}{N_1},\cdots, \frac{N}{N_c})$ may get the best prediction. It learns more equally accurately in each category. 
 \item For $L_{\rm WFCECL}$ ( $\boldsymbol{\beta}\neq {\bf 0}$ and $\boldsymbol{\alpha}\neq {\bf 0}$) , 
 its gradient is
 \begin{equation}
  \frac{\partial L}{\partial p_j}=\frac{\partial L}{\partial p_{j,+}}+ \frac{\partial L}{\partial p_{j,-}},\label{gradient}
  \end{equation}
 where
 \begin{equation}
 \frac{\partial L}{\partial p_{j,+}}= \alpha_j \frac{N_j}{N} \gamma (1-p_{j,+})^{\gamma-1}\log p_{j,+} -\alpha_j\frac{N_j}{N}\frac{(1-p_{j,+})^\gamma}{p_{j,+}}<0 
 \end{equation}
 \begin{equation}
  \frac{\partial L}{\partial p_{j,-}}=-  \beta_j \frac{N-N_j }{N}\gamma p_{j,-}^{\gamma-1} \log (1-p_{j,-})+  \beta_j \frac{N-N_j}{N} \frac{p_{j,-}^\gamma}{1-p_{j,-}}>0.
 \end{equation}
 The gradient from positive samples is negative, making the model increase $p_+$. The gradient from negative samples is positive to decrease $p_-$. Either under standard weights 
 \begin{equation*}
\begin{cases}
\begin{array}{ll}
\boldsymbol{\alpha}_s&=(\frac{N-N_1}{N}, \cdots, \frac{N-N_C}{N})\\
\boldsymbol{\beta}_s&=(\frac{N_1}{N}, \cdots, \frac{N_C}{N})
\end{array}
\end{cases},
\end{equation*}
or EE weights \eqref{eeweight}, \eqref{gradient} is written as 
 \begin{equation}
\frac{\partial L}{\partial p_j}=
\gamma (1-p_{j,+})^{\gamma-1} \log p_{j,+} -\frac{(1-p_{j,+})^\gamma}{p_{j,+}}- \gamma p_{j,-}^{\gamma-1}\log (1-p_{j,-})+\frac{p_{j,-}^\gamma}{1-p_{j,-}}\label{gradientpone}
 \end{equation}
up to scaling. Notice that  $\frac{\partial L}{\partial p_j}<0$ if $p_j<\frac12$ and $\frac{\partial L}{\partial p_j}>0$ if $p_j>\frac12$. To enlarge the range of prediction to maintain the gradient negative, we multiply a constant $\lambda\geq 1$ with $\boldsymbol{\alpha}$ in \eqref{gradientpone} . It forces the model to learn from decreasing the loss by contributing larger gradients from positive samples. That is, for $\gamma=1$, the gradient is 
\begin{equation}
\frac{\partial L}{\partial p_j}=
\lambda \log p_{j,+} -\lambda \frac{(1-p_{j,+})}{p_j}
- \log (1-p_{j,-})+\frac{p_{j,-}}{1-p_{j,-}}.\label{lambda}
 \end{equation}
If $\lambda=20$, then we get
 $\frac{\partial L}{\partial p_j}<0$ for  $p_j<0.831$. 
Table \ref{multiplication} shows that we do not need large $\lambda$ and it performs well for $18\leq \lambda\leq 24$. Experiments are done for $L_{\rm CECLA_{1,\lambda}}$ loss \eqref{awooloss} under batch=8 and 100 epochs.
\begin{table}
\centering
\begin{tabular}{ c| c| c|c|c|c|c}
\hline
$\lambda$ &$\lambda=1$&$\lambda=12$&$\lambda=18$ &$\lambda=20$ &$\lambda=22$&$\lambda=24$ \\ 
 \hline
  $\textrm{F}_1$&85.1&84.8&85.2 &85.7&85.3&85.5\\
 \hline 
 $\textrm{F}_2$&86.7&86.8&86.8&87.2&86.9&86.7\\
 \hline
  \end{tabular}
  \caption{Multiplication of $\boldsymbol{\alpha}$}
  \label{multiplication}
\end{table}
 \end{itemize}
 
  \subsection{Accuracy Comparison of Loss Functions}

\begin{table}
\centering
\begin{tabular}{c|c|c|c|c|c|c|c|c}
\hline
 model &CE&WCE&$\textrm{PBP}_{0,\textrm{s}}$&$\textrm{PBP}_{0,1}$&$\textrm{PBPA}_{0,1}$&$\textrm{CL}_{0,1}$&$\textrm{CLA}_{0,1}$&CL$_{0,\textrm{s}}$\\ 
 \hline
 category 2&7156&8631&7610&8248&8171&8733&8164&7974\\
 \hline
 category 3&3887&4426&4386&4459&4647&4386&4407&4234\\
 \hline
 \hline
 Macro ${\rm F}_1$&{\bf 79.9}&79.8&79.4&79.1&78.7&79.5&79.4&79.3\\
 \hline
 Macro ${\rm F}_2$&79.8& 83.6&81.5&82.3&82.5&82.6&83.3&81.9\\
 \hline
 \end{tabular}  
 \caption{Amount and Accuracy Comparison at 50 epochs part 2}
\label{twocategoryprediction50_1}
\end{table}

 \begin{table}
\centering
\begin{tabular}{c|c|c|c|c|c|c|c|c}
 \hline
 model &FCE&FWCE&$\textrm{PBP}_{0,8}$&$\textrm{PBPA}_{0,8}$&$\textrm{CL}_{1, 1}$&$\textrm{CLA}_{1,1}$&$\textrm{CL}_{1, 20}$&$\textrm{CLA}_{1,20}$\\ 
 \hline
 category 2&7198&9161&8375&8218&8405&8626&8885&8564\\
 \hline
 category 3&4039&4570&4406&4458&4575&4578&4572&4568\\
 \hline
 \hline
 Macro ${\rm F}_1$&79.8&79.8&{\bf 79.9}&79.5&78.6&79.4&79.1&79.4\\
 \hline
 Macro ${\rm F}_2$&80.3&79.3&83.3&82.8&82.5&{\bf 83.7}&{\bf 83.7}&83.5\\
 \hline
  \end{tabular}  
  \caption{Amount and Accuracy Comparison at 50 epochs part 2}
\label{twocategoryprediction50_2}
\end{table}

\begin{table*}
\centering
\begin{tabular}{c|c|c|c|c|c|c|c|c}
\hline
 model &CE&WCE&$\textrm{PBP}_{0,\textrm{s}}$&$\textrm{PBP}_{0,1}$&$\textrm{PBPA}_{0,1}$&$\textrm{CL}_{0,1}$&$\textrm{CLA}_{0,1}$&CL$_{0,\textrm{s}}$\\ 
 \hline
 category 2&7352&8352&8035&8049&8421&7715&7995&8111\\
 \hline
 category 3&3915&4071&3894&4133&4133&4112&4138&4287\\
 \hline
 \hline
 Macro ${\rm F}_1$&79.8&{\bf 79.9}&79.3&79.1&79.0&79.2&78.8&79.3\\
 \hline
 Macro ${\rm F}_2$&80.1&82.3&80.6&81.4&81.3&80.7&82.0&82.0\\
 \hline
 \end{tabular}  
  \caption{Amount and Accuracy Comparison at 100 Epochs Part 1}
\label{twocategoryprediction100_1}
\end{table*}
 
 \begin{table*}
\centering
\begin{tabular}{c|c|c|c|c|c|c|c|c}
 \hline
 model &FCE&FWCE&$\textrm{PBP}_{0,8}$&$\textrm{PBPA}_{0,8}$&$\textrm{CL}_{1, 1}$&$\textrm{CLA}_{1,1}$&$\textrm{CL}_{1, 20}$&$\textrm{CLA}_{1,20}$\\ 
 \hline
 category 2&7371&8031&8495&8022&8655&8148&8618&8446\\
 \hline
 category 3&3735&4111&4065&4240&4301&4296&4258&4124\\
 \hline
 \hline
 Macro ${\rm F}_1$&79.7&79.5&79.4&79.3&78.9&79.2&79.0&{\bf 79.9}\\
 \hline
 Macro ${\rm F}_2$&79.4&81.2&81.9&81.3&82.4&81.8&82.2&{\bf 82.5}\\
 \hline
  \end{tabular}  
  \caption{Amount and Accuracy Comparison at 100 Epochs Part 2}
\label{twocategoryprediction100_2}
\end{table*}

Table \ref{twocategoryprediction50_1}, \ref{twocategoryprediction50_2}, \ref{twocategoryprediction100_1} and \ref{twocategoryprediction100_2} show the accuracy comparison and category amount prediction of loss functions at 50 and 100 epochs respectively by excluding the category of none. ${\rm CL}_{1,20}$ represents WCECL loss with EE weight, focal factor $\gamma=1$ and multiplication factor $\lambda=20$. ${\rm CLA}_{1,20}$ represents WCECL loss with array weight, focal factor $\gamma=1$ and multiplication factor $\lambda=20$.${\rm PBP_{0,s}}$ notes for PBP loss with standard weight and $\gamma=0$. 

\begin{figure}[ht]
\centering
\includegraphics[width=0.8\textwidth]{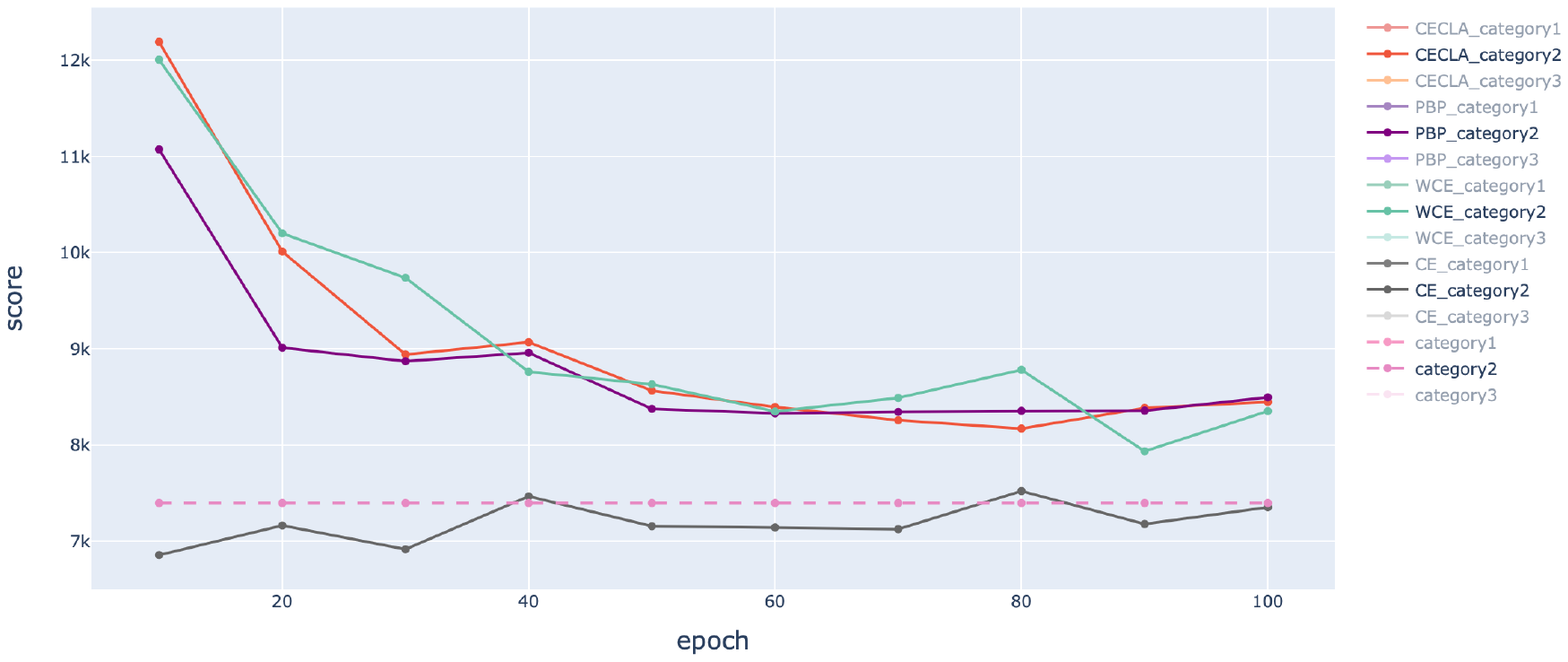} 
\caption{Amount Prediction for Category 2}
\label{countnumberprediction}
\end{figure}

$L_{\rm CE}$ and $L_{\rm PBP_{0.8}}$ both obtain the highest $\textrm{F}_1$ scores at 50 epochs. However, $L_{\rm PBP_{0.8}}$ loss has a much higher $\textrm{F}_2$. Category 2 and Category 3 represent keywords in our data. Since $L_{\rm PBP_{0.8}}$ returns more results, it is more competitive in keyword extraction. In Figure \ref{countnumberprediction}, $L_{\rm CE}$ underestimates category amount prediction. It has higher precision and lower recall (higher ${\rm F}_1$ and lower ${\rm F}_2$ shown in Table \ref{twocategoryprediction50_1},  \ref{twocategoryprediction50_2}, \ref{twocategoryprediction100_1}, and \ref{twocategoryprediction100_2}). It is not beneficial for auto-tagging. A direct computation gives that 
\[
{\rm F}_1-{\rm F} _2=\frac{3pr(p-r)}{(p+r)(4p+r)}.
\]
This formula shows the tradeoff between precision-recall and ${\rm F}_1-{\rm F}_2$. 

\begin{figure}[ht]
\centering
\includegraphics[width=0.8\textwidth]{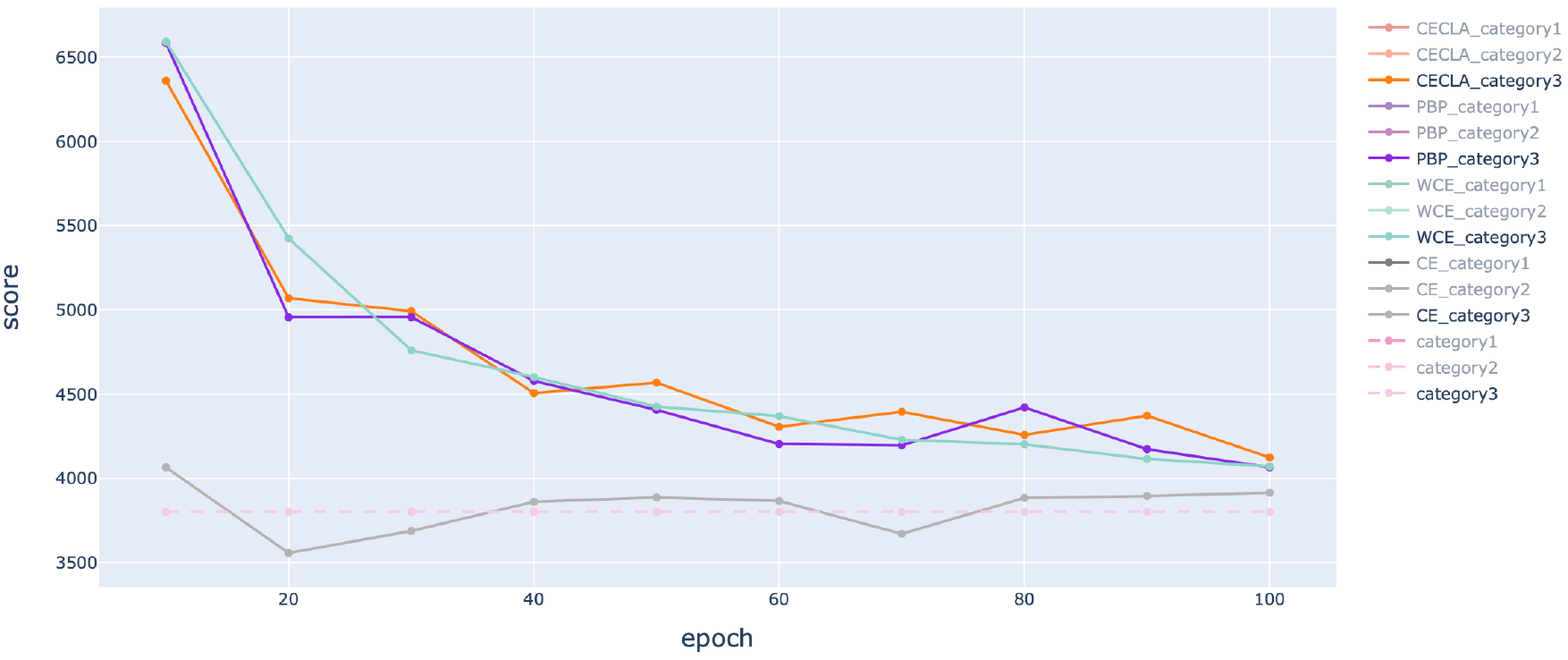}
\caption{Amount Prediction for Category 3}
\label{fig3}
\end{figure}

The dot pink horizontal line represents the true amount of category in Figure \ref{countnumberprediction} and \ref{fig3}. $L_{\rm CLA_{1,20}}$, $L_{\rm PBP_{0.8}}$, and $L_{\rm WCE}$ return more labels of keywords. Besides, they maintain almost the same ${\rm F}_1$ as $L_{\rm CE}$ but much higher ${\rm F}_2$. These new objective functions return not only more truly relevant but also more accurate results.  

$L_{\rm CECL}$ performs better with array weights at 100 epochs. $L_{\rm PBP}$ learn better with EE weights at 50 epochs.  $L_{\rm CECL}$ has more portion with EE weight than standard weight on the microdata and it has less prediction on the counts at 100 epochs. Since the negative samples of $L_{\rm CECLA}$ depend on categories, it predicts more counts than EE weight on the microdata at 100 epochs. See the following section for an explanation. 

 \subsubsection{EE Weights vs Standard Weights for CECL Loss}
 Suppose that $N_i<N_j<N-N_j<N-N_i$. That is, the data in class $i$ is more imbalanced than in class $j$. Recall the EE weights
 \begin{equation*}
\begin{cases}
\begin{array}{ll}
\boldsymbol{\alpha}_{\rm EE}&=(\frac{N}{N_1}, \cdots, \frac{N}{N_C})\\
\boldsymbol{\beta}_{\rm EE}&=(\frac{N}{(N-N_1)}, \cdots, \frac{N}{(N-N_C)})
\end{array}
\end{cases},
\end{equation*}
and standard weights
 \begin{equation*}
\begin{cases}
\begin{array}{ll}
\boldsymbol{\alpha}_{\rm st}&=(\frac{N-N_1}{N}, \cdots, \frac{N-N_C}{N})\\
\boldsymbol{\beta}_{\rm st}&=(\frac{N_1}{N}, \cdots, \frac{N_C}{N})
\end{array}
\end{cases}.
\end{equation*}

The proportion of the positive and negative weight is the same 
 \[
 (\alpha_{\rm EE,i} :\beta_{\rm EE,i})=\frac{N}{(N-N_i)N_i} (\alpha_{\rm st,i}:\beta_{\rm st,i}).
 \]
 Although the proportion $\alpha_i:\beta_i$ is the same, $\frac{N}{N_i(N-N_i)}>\frac{N}{N_j(N-N_j)}$ reveals that a single token from micro data gets more punishment than major data. The model predicts more category amounts for microdata with EE weights than standard weights in the early stage but the other around at 100 epochs. EE weights obtain less $F_1$ scores but higher $F_2$ scores.

\subsection{Comparison of WCE, CECLA, and PBP Loss}
The differences between $L_{\rm WCE}$, $L_{\rm CECLA}$, and $L_{\rm PBP}$:
\begin{itemize}
\item For contrastive learning type of functions (with $\boldsymbol{\beta}$), their gradient \eqref{gradient} has more bounce back between positive and negative samples. Its F scores also have more up and down. It has a better chance to obtain the optimal point. From Figure 6, $L_{\rm CECLA}$ achieves best macro ${\rm F}_2$ score at 40 epochs. On the other hand, $L_{\rm WCE}$ only has the contribution from positive samples. Its F scores curves are steadily increasing or decreasing. It requires a longer time to achieve the optimal. 
\item In general, $L_{\rm PBP}$ converges faster than other loss functions. It has a higher ${\rm F}_1$ than $L_{\rm WCE}$ along with training. At 40 epochs, $L_{\rm PBP}$ performs better than $L_{\rm WCE}$. 
  \end{itemize}
\begin{figure}[ht]
\centering{
\includegraphics[width=0.7\textwidth]{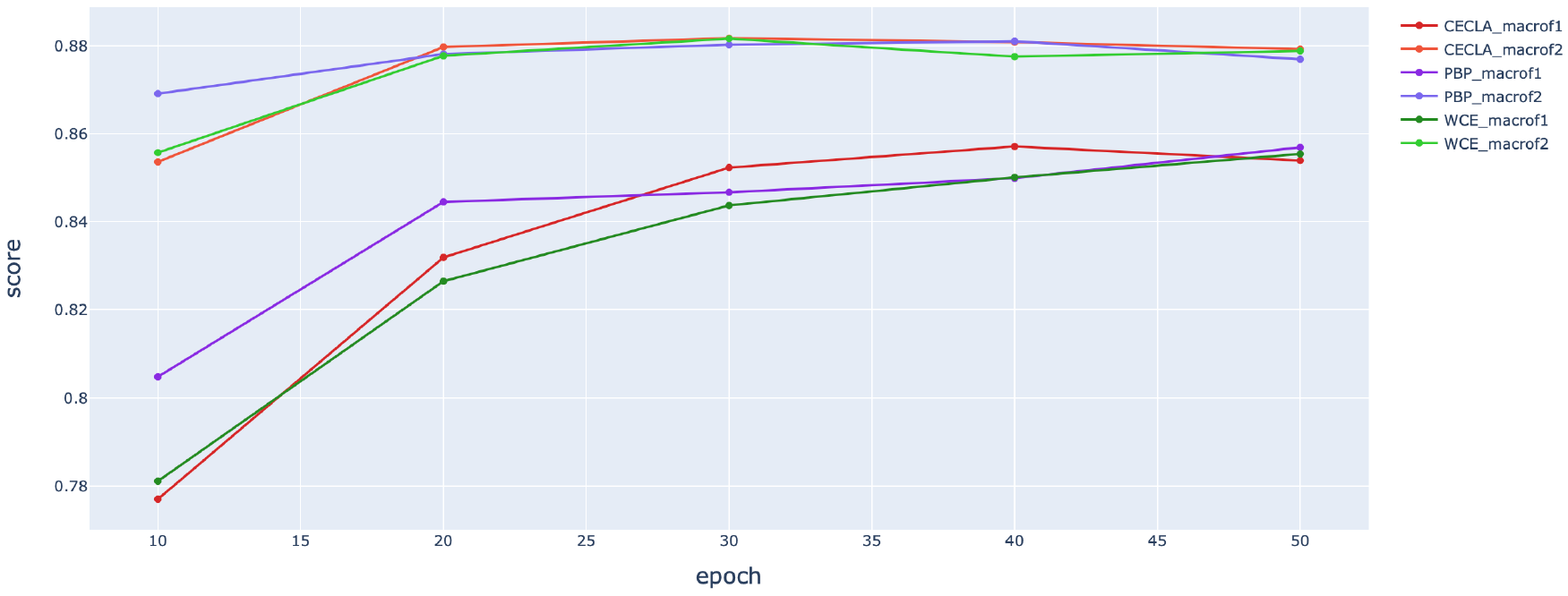} 
\label{6}
\caption{CECLA vs PBP vs WCE}
}
\end{figure}   

\section{Experiment Setup}
The source codes of SOTA models are obtained either from their official or third trustable Github. 
\begin{itemize}
\item 
Admin:\url{https://github.com/LiyuanLucasLiu/Transformer-Clinic}
\item
Muse: \url{https://github.com/lancopku/Prime}
\item 
Performers: \url{https://github.com/google-research/google-research/tree/}\\
\url{master/performer/fast_attention/tensorflow/fast_attention.py}
\item 
Transformer: \url{https://www.tensorflow.org/text/tutorials/}
\item 
Informer: \url{https://github.com/zhouhaoyi/Informer2020}
\item
Switch: \url{https://keras.io/examples/nlp/text_classification_with_switch}\\
\url{_transformer/}
\end{itemize}
\subsection{Setup for Deeper Layers}
We set ${\rm warmup\_steps}=8000$ in \eqref{lr} for 12 and 24-layer structures of all SOTA models. They adapt ${\rm lrate}/3$ for their 12-layer structure. This reduces the slope for the warmup stage smaller than it in  \eqref{lr}. 
\begin{itemize}
\item BAT changes the learning rate to ${\rm lrate}/5$ for its 24-layer structure to a finer convergence. 
\item We finalize ${\rm warmup\_steps}=12000$ and ${\rm lrate}/10$ for the 24-layer structure of Transformer. However, its convergence is not satisfying after several tries.  
\item The learning rate for the 24-layer structure of MUSE, Performers, and Informer maintains ${\rm lrate}/3$. 
\item Since the average tokens in EC-Zh is 46.6. We select the expert numbers ${\rm E}=50$ and capacity ${\rm C}=1$ for Switch. Its learning rate is converted to ${\rm lrate}/10$ for its 24-layer structure. A model with a large number of parameters needs a smaller learning rate to converge. 
\end{itemize}

\bibliographystyle{abbrv}
\bibliography{BAT}

\begin{thebibliography}{10}

\bibitem{YAKE}
R.~Campos, V.~Mangaravite, A.~Pasquali, A.~M. Jorge, C.~Nunes, and A.~Jatowt.
\newblock Yake!
\newblock In {\em Advances in Information Retrieval - 40th European Conference
  on {IR} Research, {ECIR} 2018, Grenoble, France, March 26-29, 2018,
  Proceedings}, volume 10772 of {\em Lecture Notes in Computer Science}, pages
  806--810. Springer, 2018.

\bibitem{seq2seq1}
K.~Cho, B.~van Merrienboer, D.~Bahdanau, and Y.~Bengio.
\newblock On the properties of neural machine translation: Encoder-decoder
  approaches.
\newblock In {\em Proceedings of SSST@EMNLP 2014, Eighth Workshop on Syntax,
  Semantics and Structure in Statistical Translation, Doha, Qatar, 25 October
  2014}, pages 103--111. Association for Computational Linguistics, 2014.

\bibitem{Performers}
K.~Choromanski, V.~Likhosherstov, D.~Dohan, X.~Song, A.~Gane, T.~Sarl{\'{o}}s,
  P.~Hawkins, J.~Davis, A.~Mohiuddin, L.~Kaiser, D.~Belanger, L.~Colwell, and
  A.~Weller.
\newblock Rethinking attention with performers, 2020.

\bibitem{Switch}
W.~Fedus, B.~Zoph, and N.~Shazeer.
\newblock Switch transformers: Scaling to trillion parameter models with simple
  and efficient sparsity, 2021.

\bibitem{CL}
R.~Hadsell, S.~Chopra, and Y.~LeCun.
\newblock Dimensionality reduction by learning an invariant mapping.
\newblock In {\em 2006 {IEEE} Computer Society Conference on Computer Vision
  and Pattern Recognition {(CVPR} 2006), 17-22 June 2006, New York, NY, {USA}},
  pages 1735--1742. {IEEE} Computer Society, 2006.

\bibitem{DSSM}
P.~Huang, X.~He, J.~Gao, L.~Deng, A.~Acero, and L.~P. Heck.
\newblock Learning deep structured semantic models for web search using
  clickthrough data.
\newblock In Q.~He, A.~Iyengar, W.~Nejdl, J.~Pei, and R.~Rastogi, editors, {\em
  22nd {ACM} International Conference on Information and Knowledge Management,
  CIKM'13, San Francisco, CA, USA, October 27 - November 1, 2013}, pages
  2333--2338. {ACM}, 2013.

\bibitem{Adam}
D.~P. Kingma and J.~Ba.
\newblock Adam: {A} method for stochastic optimization.
\newblock In Y.~Bengio and Y.~LeCun, editors, {\em 3rd International Conference
  on Learning Representations, {ICLR} 2015, San Diego, CA, USA, May 7-9, 2015,
  Conference Track Proceedings}, 2015.

\bibitem{Admin}
L.~Liu, X.~Liu, J.~Gao, W.~Chen, and J.~Han.
\newblock Understanding the difficulty of training transformers.
\newblock In {\em Proceedings of the 2020 Conference on Empirical Methods in
  Natural Language Processing, {EMNLP} 2020, Online, November 16-20, 2020},
  pages 5747--5763. Association for Computational Linguistics, 2020.

\bibitem{TextRank}
R.~Mihalcea and P.~Tarau.
\newblock {TextRank}: Bringing order into texts.
\newblock In {\em Proceedings of {EMNLP-04}and the 2004 Conference on Empirical
  Methods in Natural Language Processing}, July 2004.

\bibitem{reranker}
R.~Nogueira and K.~Cho.
\newblock Passage re-ranking with {BERT}, 2019.

\bibitem{GPT2}
A.~Radford, K.~Narasimhan, T.~Salimans, and I.~Sutskever.
\newblock Improving language understanding with unsupervised learning.
\newblock https://openai.com/blog/language-unsupervised/, 2018.

\bibitem{RAKE}
S.~Rose, D.~Engel, N.~Cramer, and W.~Cowley.
\newblock Automatic keyword extraction from individual documents.
\newblock In M.~W. Berry and J.~Kogan, editors, {\em Text Mining. Applications
  and Theory}, pages 1--20. John Wiley and Sons, Ltd, 2010.

\bibitem{facenet}
F.~Schroff, D.~Kalenichenko, and J.~Philbin.
\newblock Facenet: A unified embedding for face recognition and clustering.
\newblock In {\em {IEEE} Conference on Computer Vision and Pattern Recognition,
  {CVPR} 2015, Boston, MA, USA, June 7-12, 2015}, pages 815--823. {IEEE}
  Computer Society, 2015.

\bibitem{ET}
D.~R. So, C.~Liang, and Q.~V. Le.
\newblock The evolved transformer.
\newblock In {\em Proceedings of the 36th International Conference on Machine
  Learning, {ICML} 2019, 9-15 June 2019, Long Beach, California, {USA}},
  volume~97 of {\em Proceedings of Machine Learning Research}, pages
  5877--5886. {PMLR}, 2019.

\bibitem{MCL}
K.~Sohn.
\newblock Improved deep metric learning with multi-class n-pair loss objective.
\newblock In {\em Advances in Neural Information Processing Systems 29: Annual
  Conference on Neural Information Processing Systems 2016, December 5-10,
  2016, Barcelona, Spain}, pages 1849--1857, 2016.

\bibitem{keyBERT}
M.~Tang, P.~Gandhi, M.~A. Kabir, C.~Zou, J.~Blakey, and X.~Luo.
\newblock Progress notes classification and keyword extraction using
  attention-based deep learning models with {BERT}, 2019.

\bibitem{Transformer}
A.~Vaswani, N.~Shazeer, N.~Parmar, J.~Uszkoreit, L.~Jones, A.~N. Gomez,
  L.~Kaiser, and I.~Polosukhin.
\newblock Attention is all you need.
\newblock In {\em Advances in Neural Information Processing Systems},
  volume~30. Curran Associates, Inc., 2017.

\bibitem{singlerank}
X.~Wan and J.~Xiao.
\newblock Collabrank: Towards a collaborative approach to single-document
  keyphrase extraction.
\newblock In D.~Scott and H.~Uszkoreit, editors, {\em {COLING} 2008, 22nd
  International Conference on Computational Linguistics, Proceedings of the
  Conference, 18-22 August 2008, Manchester, {UK}}, pages 969--976, 2008.

\bibitem{KNRM}
C.~Xiong, Z.~Dai, J.~Callan, Z.~Liu, and R.~Power.
\newblock End-to-end neural ad-hoc ranking with kernel pooling.
\newblock In {\em Proceedings of the 40th International {ACM} {SIGIR}
  Conference on Research and Development in Information Retrieval, Shinjuku,
  Tokyo, Japan, August 7-11, 2017}, pages 55--64. {ACM}, 2017.

\bibitem{DRMM}
Z.~Yang, Q.~Lan, J.~Guo, Y.~Fan, X.~Zhu, Y.~Lan, Y.~Wang, and X.~Cheng.
\newblock A deep top-k relevance matching model for ad-hoc retrieval.
\newblock In {\em Information Retrieval - 24th China Conference, {CCIR} 2018,
  Guilin, China, September 27-29, 2018, Proceedings}, volume 11168 of {\em
  Lecture Notes in Computer Science}, pages 16--27. Springer, 2018.

\bibitem{keyBLSTM}
Q.~Zhang, Y.~Wang, Y.~Gong, and X.~Huang.
\newblock Keyphrase extraction using deep recurrent neural networks on twitter.
\newblock In J.~Su, X.~Carreras, and K.~Duh, editors, {\em Proceedings of the
  2016 Conference on Empirical Methods in Natural Language Processing, {EMNLP}
  2016, Austin, Texas, USA, November 1-4, 2016}, pages 836--845. The
  Association for Computational Linguistics, 2016.

\bibitem{MUSE}
G.~Zhao, X.~Sun, J.~Xu, Z.~Zhang, and L.~Luo.
\newblock {MUSE:} parallel multi-scale attention for sequence to sequence
  learning, 2019.

\bibitem{Informer}
H.~Zhou, S.~Zhang, J.~Peng, S.~Zhang, J.~Li, H.~Xiong, and W.~Zhang.
\newblock Informer: Beyond efficient transformer for long sequence time-series
  forecasting, 2020.

\end{thebibliography}
\end{document}